\documentclass{article}
\usepackage{graphicx} 
\usepackage{lipsum} 
\usepackage{authblk}
\usepackage{datetime}
\usepackage[multiple]{footmisc}
\usepackage{makecell}
\usepackage{array}
\usepackage{url}
\usepackage[font=small,labelfont=bf]{caption}
\usepackage{float}
\usepackage[numbers]{natbib}
\newcommand{\minus}{\scalebox{0.75}[1.0]{$-$}}
\usepackage{neurips_data_2024}

\begin{document}

\title{That Chip Has Sailed: A Critique of Unfounded Skepticism Around AI for Chip Design}
\author[1,3]{Anna Goldie}
\author[1,3]{Azalia Mirhoseini}
\author[1,2]{Jeff Dean}
\affil[1]{Google DeepMind}
\affil[2]{Google Research}
\affil[3]{Department of Computer Science, Stanford University}
\date{\today}

\maketitle

\begin{abstract}

In 2020, we introduced a deep reinforcement learning method capable of generating superhuman chip layouts, which we then published in \emph{Nature} and open-sourced on GitHub. AlphaChip has inspired an explosion of work on AI for chip design, and has been deployed in state-of-the-art chips across Alphabet and extended by external chipmakers. Even so, a non-peer-reviewed invited paper at ISPD 2023 questioned its performance claims, despite failing to run our method as described in \emph{Nature}. For example, it did not pre-train the RL method (removing its ability to learn from prior experience), used substantially fewer compute resources (20x fewer RL experience collectors and half as many GPUs), did not train to convergence (standard practice in machine learning), and evaluated on test cases that are not representative of modern chips. Recently, Igor Markov published a ``meta-analysis'' of three papers: our peer-reviewed \emph{Nature} paper, the non-peer-reviewed ISPD paper, and Markov's own unpublished paper (though he does not disclose that he co-authored it). Although AlphaChip has already achieved widespread adoption and impact, we publish this response to ensure that no one is wrongly discouraged from innovating in this impactful area.
\end{abstract}

\section{Introduction}

\begin{figure}[hbt!]
    \centering
    \includegraphics[width=\textwidth]{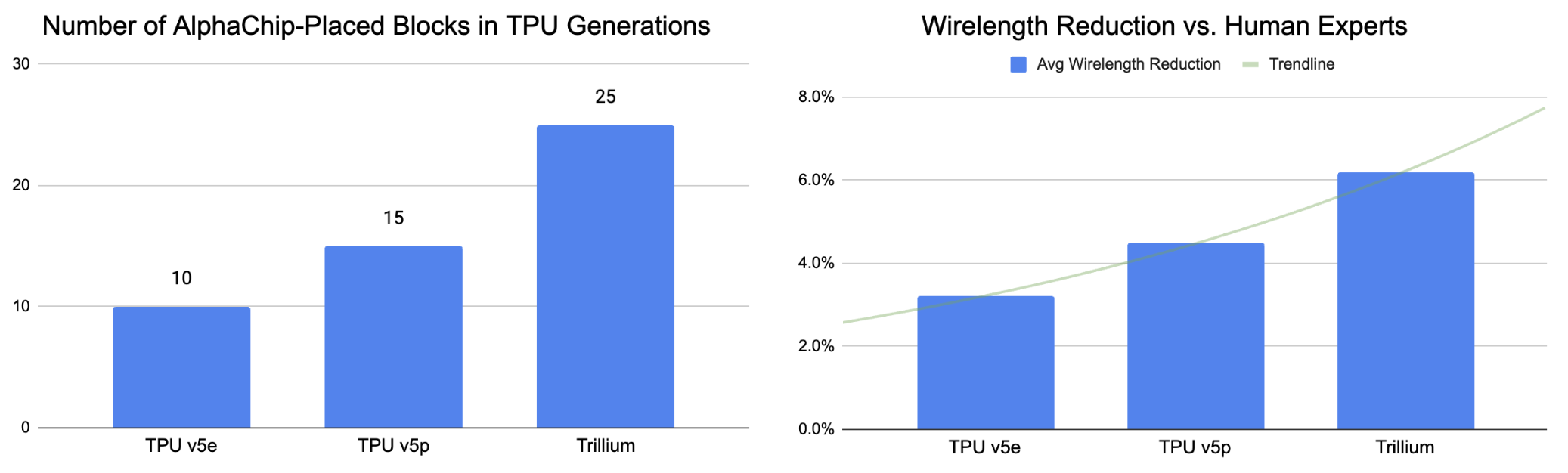}
    
    \caption{AlphaChip has been deployed in three additional generations of TPU. In each generation, it has been adopted in a greater proportion of blocks and has outperformed human experts by a wider margin.}
    \label{fig:alphachip-adoption}
\end{figure}

Following its publication in \emph{Nature}, AlphaChip \citep{mirhoseini-goldie2021} has inspired an explosion of work on AI for chip design \citep{lu2023dac,lai2023chipformer, shi2024wiremask, lai2022maskplace, lai2021maskplace, chang2023flexible, yan2024human, synopsys2024, cadence2024, chuang2024aws, forbes2023, eetimes2023, ieee2023, cadence2022}. It has also generated superhuman chip layouts used in three generations of TPU (see Figure \ref{fig:alphachip-adoption}), datacenter CPUs (Axion), and other chips across Alphabet, and been extended to new areas of chip design by external academics and chipmakers \cite{mediatek2022,chowdhuryrlsynthesis2024}.

Even so, Igor Markov published a criticism of our work in the November 2024 issue of Communications of the ACM \citep{cacm2024}, which is presented as a ``meta-analysis'' of our \emph{Nature} paper and two non-peer-reviewed papers:

\begin{itemize}
    \item \textbf{Cheng \emph{et al.}}: The first is an invited ISPD paper\footnote{Invited papers at ISPD are not peer-reviewed.} by Cheng \emph{et al.} \cite{cheng2023assessment}. This paper did not follow standard machine learning practices, and its reinforcement learning methodology and experimental setup diverged significantly from those described in our \emph{Nature} paper. Nevertheless, its hamstrung version of our method still outperformed RePlAce\footnote{Incidentally, RePlAce, as noted in a footnote of Cheng \emph{et al.}, is unable to produce any result at all for 2 out of the 6 test cases in its main data table.} \citep{cheng2019replace}, which was the state of the art when we published in \emph{Nature}.
    \item \textbf{Markov \emph{et al.}}: The second ``meta-analyzed'' paper is an unpublished PDF with no author list \citep{markovetal}, which is described as a ``separate evaluation'' performed by ``Google Team 2'', but was in fact co-authored by Markov himself\footnote{Markov did not disclose anywhere in his ``meta-analysis'' that he is an author of one of the two ``separate evaluations''. He also omitted his name from the paper's authors in the references section, and linked only to an anonymous PDF. When questioned on LinkedIn, Markov admitted his authorship, but later deleted the post.}, though this is not disclosed\footnote{Markov also failed to disclose his role as a high-level employee at Synopsys, a company which licenses commercial tools that compete with our open-source method.}. This paper did not meet Google's bar for publication. In 2022, it was reviewed by an independent committee at Google, which determined that ``the claims and conclusions in the draft are not scientifically backed by the experiments'' \citep{orwant2022} and ``as the [AlphaChip] results on their original datasets were independently reproduced, this brought the [Markov \emph{et al.}] RL results into question'' \citep{orwant2022}. We provided the committee with one-line scripts that generated significantly better RL results than those reported in Markov \emph{et al.}, outperforming their ``stronger'' simulated annealing baseline. We still do not know how Markov and his collaborators produced the numbers in their paper.
\end{itemize}

Markov's ``meta-analysis'' offers one additional source of concern regarding our paper: a ``whistleblower'' within Google. However, this ``whistleblower'' admitted to a Google investigator that he had no reason to believe fraud occurred: ``he stated that he suspected that the research being conducted by Goldie and Mirhoseini was fraudulent, but also stated that he did not have evidence to support his suspicion of fraud'' \citep{laster2022}.

In his ``meta-analysis'', Markov speculates wildly and without evidence about fraud and scientific misconduct, none of which occurred. Most of Markov's criticisms are of this form: it does not look to him like our method \emph{should} work, and therefore it \emph{must not} work, and any evidence suggesting otherwise is fraud. \emph{Nature} investigated Markov's concerns, found them to be entirely without merit, and published an Addendum upholding our work at the conclusion of this process \citep{goldie-mirhoseini2024}.

As an example, in the opening paragraph of his conclusions, Markov states (emphasis his): ``In the paper, we find a smorgasbord of \emph{questionable practices in ML} [26]\footnote{Note that Markov's citation 26 has nothing to do with our paper, though readers may mistakenly believe that it offers corroboration.} including irreproducible research practices, multiple variants of cherry-picking, misreporting, and likely data contamination (leakage).'' We did not engage in any of these practices, or any other form of scientific misconduct, and Markov provides no evidence for these allegations. Nowhere in Markov’s paper does he describe any form of alleged cherry-picking, let alone multiple variants, nor does he provide evidence. Nor does he describe any form of alleged ``misreporting,'' or explain what he means by this, or provide evidence. Nor does he provide any evidence of data contamination (leakage), aside from his speculation that it would have improved our results if it had occurred. Many of these allegations appear \emph{for the first time} in his ``Conclusions'' section!

In an effort to discredit our TPU deployments, Markov also suggests that Google must just be ``dogfooding'' our method, allowing inferior AlphaChip placements to be used in TPU in order to prop up our research paper. This is untrue, and absurd on its face. Google cares far more about the efficiency of TPU designs -- a multi-billion-dollar project that is central to Google's cloud and AI initiatives -- than it does about a research paper\footnote{In reality, we had to work for a long time to build enough trust for the TPU team to use our layouts, even after AlphaChip was outperforming human experts on the metrics, and this makes sense -- their job is to deliver TPU chips and make them as efficient and reliable as possible, and they cannot afford to take unnecessary risks.}.

For clarity, we present a timeline of events, including non-confidential deployments\footnote{AlphaChip has been deployed in other hardware across Alphabet that we cannot yet disclose.}:

\begin{itemize}
    \item \textbf{Apr 2020}: Released arXiv preprint of our \emph{Nature} paper \citep{preprint2020}.
    \item \textbf{Aug 2020}: 10 AlphaChip layouts taped out in TPU v5e.
    \item \textbf{Jun 2021}: Published \emph{Nature} article \citep{mirhoseini-goldie2021}.
    \item \textbf{Sep 2021}: 15 AlphaChip layouts taped out in TPU v5p.
    \item \textbf{Jan 2022 - Jul 2022}: Open-sourced AlphaChip \citep{guadarrama2021circuittraining}, after ensuring compliance with export control restrictions and excising internal dependencies. This involved independent replication of the results in our \emph{Nature} paper by another team at Google. See Section \ref{transparency}. 
    \item \textbf{Feb 2022}: Independent committee within Google declined to publish Markov \emph{et al.} as the data did not support its claims and conclusions \citep{orwant2022}.
    \item \textbf{Oct 2022}: 25 AlphaChip layouts taped out in Trillium (latest public TPU).
    \item \textbf{Feb 2023}: Cheng \emph{et al.} posted on arXiv \citep{cheng2023assessment}, claiming to perform ``massive reimplementation'' of our method, despite it being fully open-source. As discussed in Sections \ref{errors-in-reproduction} and \ref{other-issues}, Cheng \emph{et al.} did not run our method as described in \emph{Nature}, among other issues.
    \item \textbf{Jun 2023}: Markov released arXiv preprint of his ``meta-analysis'' \citep{markovarxiv2023}.
    \item \textbf{Sep 2023}: \emph{Nature} posted Editor's note stating that they are investigating our paper, and initiated second peer review process \citep{mirhoseini-goldie2021}.
    \item \textbf{Mar 2024}: 7 AlphaChip layouts adopted in Google Axion Processors (ARM-based CPU).
    \item \textbf{Apr 2024}: \emph{Nature} completed its investigation and post-publication review, and found entirely in our favor, concluding that ``the best way forward is to publish an update to the paper in the form of an Addendum (not a ‘Correction’, as we have established that there is little that actually needs correcting).'' \citep{ziemelis2024}
    \item \textbf{Sep 2024}: \emph{Nature} published Addendum upholding our work \citep{goldie-mirhoseini2024}, removed Editor's note.
    \item \textbf{Sep 2024}: SVP of MediaTek announced that they extended AlphaChip to accelerate development of their most advanced chips \citep{alphachipblog2024}.
    \item \textbf{Nov 2024}: Markov republished his ``meta-analysis'', though his concerns were already found to be without merit during \emph{Nature}'s investigation and second peer review process.
\end{itemize}

In brief, Markov's paper contains no original data, and is a ``meta-analysis'' of just two papers. The first is presented with no author list (though Markov was an author), was never published, made claims that were not scientifically backed by the data, and could not be reproduced. The second, Cheng \emph{et al.}, is the only substantive content in Markov's ``meta-analysis'', so we devote the remainder of this paper to describing significant issues in its purported reproduction of our method.

\section{Errors In Cheng \emph{et al.}'s Reproduction of Our Method}
\label{errors-in-reproduction}

Cheng \emph{et al.} claim to evaluate our method against alternative approaches on new test cases. Unfortunately, Cheng \emph{et al.} did not run our method as described in \emph{Nature}, so it is unsurprising that they report different results. In this section, we describe major errors in their purported reproduction:

\begin{itemize}
    \item \textbf{Did not pre-train the RL method.} The ability to learn from prior experience is the key advantage of our learning-based method, and to remove it is to evaluate a different and inferior approach. Incidentally, pre-training also gives rise to the impressive capabilities of large language models like Gemini \citep{google2023gemini} and ChatGPT \citep{openai2024chatgpt} (the ``P'' in ``GPT'' stands for ``pre-trained''). See Section \ref{pre-training}.
    \item \textbf{Used an order of magnitude fewer compute resources}: 20x fewer RL experience collectors (26 vs 512 in \emph{Nature}) and 2x fewer GPUs (8 vs 16 in \emph{Nature}). See Section \ref{compute}.
    \item \textbf{Did not train to convergence}. Training to convergence is standard practice in machine learning, as not doing so is well known to harm performance \citep{ml101}. See Section \ref{convergence}.
    \item \textbf{Evaluated on non-representative, irreproducible benchmarks}. Cheng \emph{et al.}'s benchmarks have much older and larger technology node sizes (45nm and 12nm vs sub-7nm in \emph{Nature}), and differ substantially from a physical design perspective. Additionally, the authors were unable or unwilling to share the synthesized netlists necessary to replicate the results in their main data table. See Sections \ref{benchmarks} and \ref{refused}.
    \item \textbf{Performed ``massive reimplementation'' of our method}, which may have introduced errors. We recommend instead using our open-source code. See Section \ref{transparency}.
\end{itemize}

These major methodological differences unfortunately invalidate Cheng \emph{et al.}'s comparisons with and conclusions about our method. If Cheng \emph{et al.} had reached out to the corresponding authors of the \emph{Nature} paper\footnote{Prior to publication of Cheng \emph{et al.}, our last correspondence with any of its authors was in August of 2022 when we reached out to share our new contact information.}, we would have gladly helped them to correct these issues prior to publication\footnote{In contrast, prior to publishing in \emph{Nature}, we corresponded extensively with Andrew Kahng, senior author of Cheng \emph{et al.} and of the prior state of the art (RePlAce), to ensure that we were using the appropriate settings for RePlAce.}.

\subsection{No Pre-Training Performed for RL Method}
\label{pre-training}

Unlike prior approaches, AlphaChip is a learning-based method, meaning that it becomes better and faster as it solves more instances of the chip placement problem. This is achieved by pre-training, which consists of training on ``practice'' blocks (training data) prior to running on the held-out test cases (test data).

\begin{figure}[hbt!]
    \centering
    \includegraphics[width=.65\textwidth]{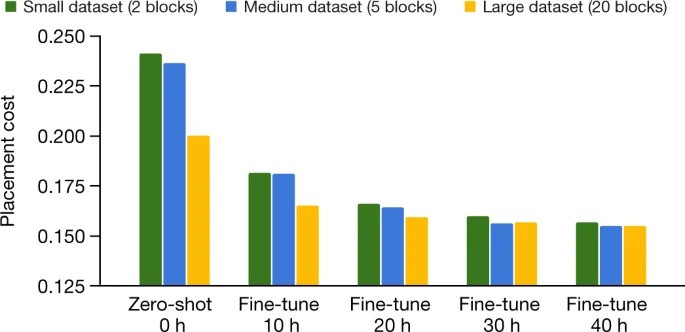}
    
    \caption{Figure 5 of the \emph{Nature} paper (reproduced above) shows performance gains from pre-training on a larger number of blocks. As we scale up the pre-training dataset size, the RL agent's performance improves.}
    \label{fig:nature-figure5}
\end{figure}

As we showed in Figure 5 of our \emph{Nature} paper (reproduced below as Figure \ref{fig:nature-figure5}), the larger the training dataset is, the better the method becomes at placing new blocks. As described in our \emph{Nature} article, we pre-trained on 20 blocks in our main data table (\emph{Nature} Table 1).

Cheng \emph{et al.} did not pre-train at all (i.e., no training data), meaning that the RL agent had never seen a chip before and had to learn how to perform placement from scratch for each of the test cases. This \textbf{removed the key advantage of our method}, namely its ability to learn from prior experience.

By analogy to other well-known work on reinforcement learning, this would be like evaluating a version of AlphaGo \citep{alphago} that had never seen a game of Go before (instead of being pre-trained on millions of games), and then concluding that AlphaGo is not very good at Go.

\begin{figure}[hbt!]
    \centering
    \includegraphics[width=.6\textwidth]{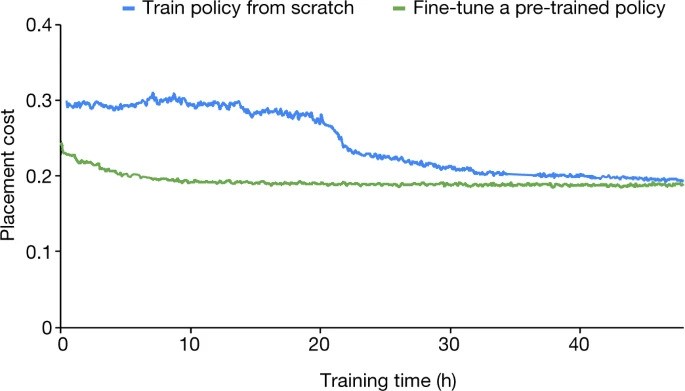}
    
    \caption{Figure 4 of the \emph{Nature} paper (reproduced above) showed that pre-training improves convergence speed compared to starting from a randomly initialized policy. On the open-source Ariane RISC-V CPU, the randomly initialized policy took 48 hours to approach what the pre-trained policy could produce in 6 hours.}
    \label{fig:nature-figure4}
\end{figure}

We discussed the importance of pre-training at length in our \emph{Nature} paper (e.g. the word ``pre-train'' appeared 37 times), and empirically demonstrated its impact. For example, \emph{Nature} Figure 4 (reproduced here as Figure \ref{fig:nature-figure4}) showed that pre-training improves placement quality and convergence speed. On the open-source Ariane RISC-V CPU \citep{zaruba2019ariane}, it took 48 hours for the non-pretrained RL policy to approach what the pre-trained model could produce in 6 hours. As described in our \emph{Nature} paper, we pre-trained for 48 hours for the results in our main data table, whereas Cheng \emph{et al.} pre-trained for 0 hours.

Our open-source repository \citep{guadarrama2021circuittraining} enables full reproduction of the methods described in our \emph{Nature} paper (see Section \ref{transparency}). Cheng \emph{et al.} have attempted to excuse their lack of pre-training by suggesting that our open-source repository does not support pre-training \citep{cheng2023assessment}, but this is incorrect. Pre-training
is simply running the method on multiple examples, and this has been always been supported.

\subsection{RL Method Provided with Far Fewer Compute Resources}
\label{compute}

In Cheng \emph{et al.}, the RL method is provided with 20x fewer RL experience collectors (26 vs 512 in \emph{Nature}) and half as many GPUs (8 vs 16 in \emph{Nature}). Using less compute is likely to harm performance, or require running for considerably longer to achieve the same (or worse) performance.

\begin{figure}[hbt!]
    \centering
    \includegraphics[width=\textwidth]{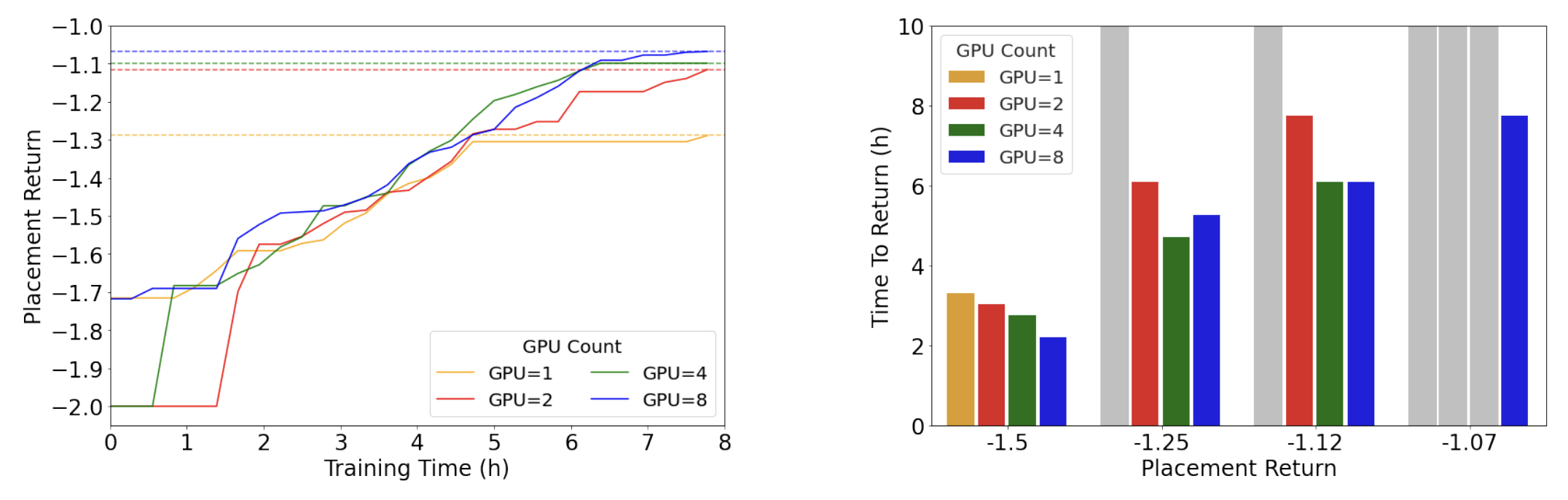}
    
    \caption{Figure 6 from a follow-up paper \citep{yue2022circuittraining} (reproduced above) demonstrated that speed and quality improve with additional compute resources. Left: Placement return (higher is better) vs. training time as a function of the number of GPUs. An infeasible placement receives a \minus 2 placement return. Increasing the number of GPUs results in better final placements. Right: Time to reach a given placement return as a function of the number of GPUs. The grey bars indicate that the experiment did not reach a specific return value. The best placement return \minus 1.07 can only be achieved with GPU=8, the largest setting in this experiment.}
    \label{fig:scalability}
\end{figure}

As shown in Figure \ref{fig:scalability} (reproduced from a follow-up paper \citep{yue2022circuittraining}), training on a larger number of GPUs speeds convergence and yields better final quality. If Cheng \emph{et al.} had matched the experimental settings described in \emph{Nature}, this would likely have improved their results. 

\subsection{RL Method Not Trained to Convergence}
\label{convergence}

As a machine learning model trains, loss typically decreases and then plateaus, representing ``convergence'' — the model has learned what it can about the task it is performing. Training to convergence is standard practice in machine learning, and not doing so is well known to harm performance \citep{ml101}. 

Cheng \emph{et al.} \textbf{did not train to convergence on any of the four blocks for which convergence plots were provided} on their accompanying project site \citep{kahng2022github} (no plots were provided for BlackParrot-NG45 or Ariane-NG45).

\begin{figure}[hbt!]
    \centering
    \includegraphics[width=\textwidth]{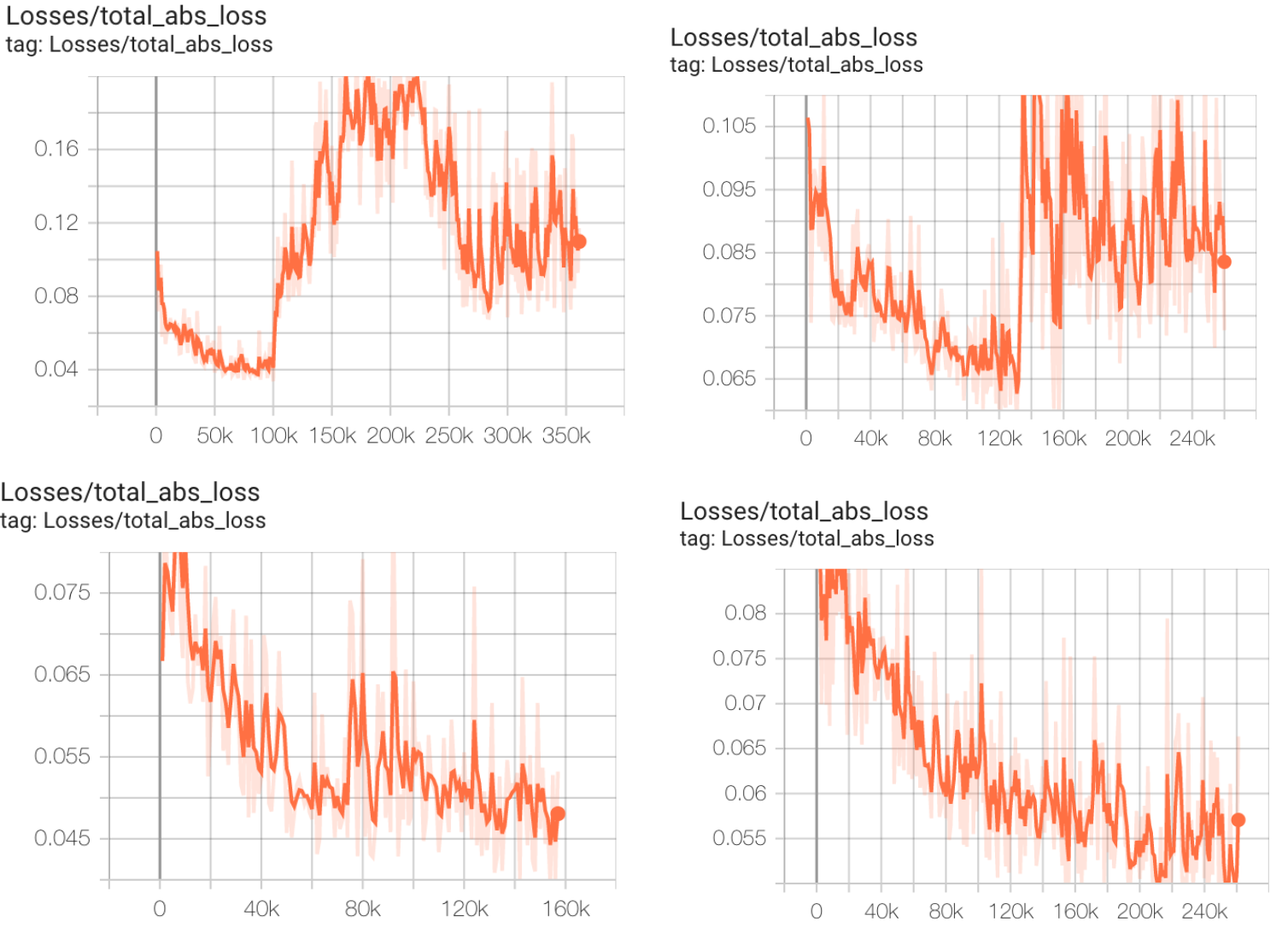}
    
    \caption{Convergence plots from Cheng \emph{et al.}'s project site. On Ariane-NG45 (top left) and MemPool-NG45 (top right), there is an odd divergence at around 100k steps, but loss appears to be trending downwards and would likely have improved with further training. On BlackParrot-GF12 (bottom left) and MemPool-GF12 (bottom right), the model has not yet converged and would likely benefit from additional training time as well.
}
    \label{fig:mempool-blackparrot}
\end{figure}

Figure \ref{fig:mempool-blackparrot} shows the convergence plots from Cheng \emph{et al.}'s project site, and Table \ref{tab:convergence-table} summarizes the information available. For all four blocks with convergence plots (Ariane-GF12, MemPool-NG45, BlackParrot-GF12, and MemPool-GF12), training was cut off at a relatively low step count (350k, 250k, 160k, and 250k steps, respectively)\footnote{Although Cheng \emph{et al.}'s Figure 4 appears to show convergence on Ariane-NG45 after 1M steps, it omits most components of the total training loss, depicting only wirelength, density, and congestion costs. However, total loss is composed of entropy regularization loss, KL penalty loss, L2 regularization loss, policy gradient loss, and value estimate loss. See open-source code for details of training loss: \url{https://github.com/google-research/circuit_training/blob/90fbe0e939c3038e43db63d2cf1ff570e525547a/circuit_training/learning/agent.py#L408}. Cheng \emph{et al.} did not provide the Tensorboard for this block, and as shown in Table \ref{tab:convergence-table}, all other blocks were run for far fewer than 1M steps.}. Following standard machine learning practices would likely improve performance on these test cases.

\begin{table}[hbt!]
\centering
\renewcommand{\arraystretch}{1.25} 
\setlength{\tabcolsep}{6pt} 
\normalsize 
\begin{tabular}{|l|c|c|p{3.75cm}|}
\hline
Block Name from Cheng \emph{et al.}'s Table 1 & Tensorboard? & Num Steps & Total Loss Curve \\ \hline
Ariane-NG45                           & No                    & 1M                 & No Tensorboard. \\ \hline
BlackParrot-NG45                       & No                    & ?                  & No Tensorboard. \\ \hline
MemPool-NG45                           & Yes                   & 250k               & Divergence at 100k steps, clearly has not converged. \\ \hline
Ariane-GF12                            & Yes                   & 350k               & Divergence at 130k steps, clearly has not converged. \\ \hline
BlackParrot-GF12                       & Yes                   & 160k               & Still converging, training stopped prematurely. \\ \hline
MemPool-GF12                           & Yes                   & 250k               & Still converging, training stopped prematurely. \\ \hline
\end{tabular}
\vspace{0.25cm} 
\caption{Cheng \emph{et al.} did not train properly on any of the test cases for which Tensorboards were provided on the accompanying project site.}
\label{tab:convergence-table}
\vspace{-0.5cm} 
\end{table}

\subsection{Cheng \emph{et al.}'s Test Cases Not Representative of Modern Chips}
\label{benchmarks}

In our \emph{Nature} paper, we report results on Tensor Processing Unit (TPU) blocks with sub-7nm technology node size, which is typical of modern chips. In contrast, Cheng \emph{et al.} reports results on older technology node sizes (45nm and 12nm), which differ substantially from a physical design perspective; for example, at sub-10nm, multiple patterning is typically used \citep{pan2015,multiplepatterning2024}, causing routing congestion issues to emerge at lower density. Therefore, for older technology node sizes, our method may benefit from adjustment to the congestion or density components of its reward function\footnote{Google engineers suggested this, but their guidance was not followed (see Section \ref{false-validation}).}. We have not focused on applying our technique to designs with older nodes because all of our work is at 7nm, 5nm, and more recent processes, though we would welcome contributions from the community on this front.

\section{Other Issues With Cheng \emph{et al.}}
\label{other-issues}

In this section, we describe other concerns with Cheng \emph{et al.}, including its comparison with closed-source commercial autoplacers, its contrived ``ablation'' of initial placement in standard cell cluster rebalancing, its flawed correlation study, and its erroneous claim of validation by Google engineers.

\subsection{Inappropriate Comparison With Commercial Autoplacers}

Cheng \emph{et al.} compares a severely weakened RL method against unpublished, closed-source, proprietary software released years after our method was published. This is not a reasonable way to evaluate our method -- for all we know, the closed-source tool could have built directly on our work.

In May of 2020, we performed a blind internal study\footnote{Our blind study compared RL to human experts and commercial autoplacers on 20 TPU blocks. First, the physical design engineer responsible for placing a given block ranked anonymized placements from each of the competing methods, evaluating purely on final QoR metrics with no knowledge of which method was used to generate each placement. Next, a panel of seven physical design experts reviewed each of the rankings and ties. The comparisons were unblinded only after completing both rounds of evaluation. The result was that the best placement was produced most often by RL, followed by human experts, followed by commercial autoplacers.} comparing our method against the latest version of two leading commercial autoplacers. Our method outperformed both, beating one 13 to 4 (with 3 ties) and the other 15 to 1 (with 4 ties). Unfortunately, standard licensing agreements with commercial vendors prohibit public comparison with their offerings.

\subsection{Contrived ``Ablation'' of Initial Placement in Standard Cell Cluster Rebalancing}
\label{initial-placement}

Prior to running the methods evaluated in our \emph{Nature} paper, an approximate initial placement from physical synthesis, the previous step of the chip design
process \citep{physicalsynthesis2024}, was used to resolve imbalances in the sizes of standard cell clusters from
hMETIS \citep{karypis1998hmetis}. 

Cheng \emph{et al.} ran an ``ablation'' study on a single block (Ariane-NG45). Instead simply skipping the cluster rebalancing step, they tried placing all chip components on top of each other in the lower-left corner\footnote{Cheng \emph{et al.} also tried placing all components on top of each other in the upper-right corner and on a single point in the center of the canvas. Unsurprisingly, this yielded the same degenerate results.}, causing the rebalancing step to produce degenerate standard cell clusters. When this harmed performance, Cheng \emph{et al.} concluded that our RL agent was somehow making use of initial placement information, even though it does not have access to the initial placement and does not place standard cells.

We ran an ablation study which eliminated any use of initial placement whatsoever, and observed no degradation in performance (see Table \ref{tab:cluster-ablation}). We simply skipped the cluster rebalancing step and instead reduced hMETIS’s cluster ``unbalancedness'' parameter to its lowest setting (UBFactor=1)\footnote{UBfactor is a parameter that ranges from 1 to 49, where lower settings instruct hMETIS to prioritize balanced cluster sizes. UBfactor was set to 5 in our \emph{Nature} paper.}, which causes hMETIS to generate more balanced clusters \citep{karypis1998hmetis}. This ancillary preprocessing step has been documented and open-sourced since June 10, 2022 \citep{guadarrama2021circuittraining}, but is unnecessary and has already been removed from our production workflow.

\begin{table}[hbt!]
\renewcommand{\arraystretch}{1.2}
\begin{tabular}{|p{2.575cm}|c|c|c|c|c|c|}
\hline
TPU-v6 Block & wirelength & wns & tns & density & congestion (H) & congestion (V) \\
\hline
Clustering with initial placement & 5,176 & -0.046 & -2.466 & 23.830 & 0.01 & 0.01 \\
\hline
Clustering with no initial placement & 5,133 & -0.048 & -2.583 & 23.827 & 0.01 & 0.01 \\
\hline
\end{tabular}\vspace{0.25cm}
\captionof{table}{RL results after clustering standard cells with and without the initial placement. Lower magnitude is better for all metrics. Clustering without initial placement does not appear to harm performance.}\label{tab:cluster-ablation}
\end{table}

\subsection{Flawed Study of Correlation Between Proxy Cost and Final Metrics}
\label{correlation-study}

\begin{figure}[hbt!]
    \centering
    \includegraphics[width=.5\textwidth]{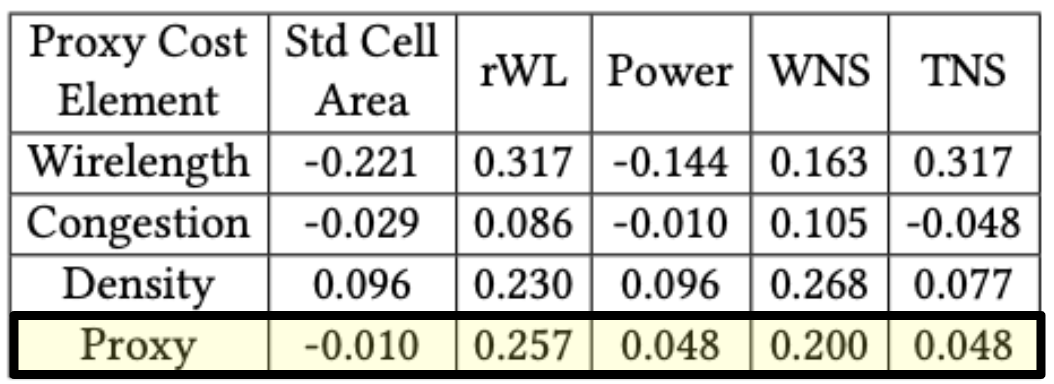}
    
    \caption{Cheng \emph{et al.}'s Table 2 (reproduced above, emphasis ours) showed a weak but positive correlation between overall proxy cost and final metrics, except std cell area, which we treat as a hard constraint and do not optimize.}
    \label{fig:ispd-table2}
\end{figure}

Cheng \emph{et al.} claimed that our proxy costs are not well-correlated with final metrics, but their correlation study actually showed a weak but positive correlation between overall proxy cost and all final metrics except standard cell area (see Cheng \emph{et al.}'s Table 2, reproduced here as Figure \ref{fig:ispd-table2}). Note that we treat area as a hard constraint, and therefore do not optimize for it.

Proxy costs used in ML-based optimization are often only weakly correlated with the target objective. For example, large language models like Gemini \citep{google2023gemini} and  ChatGPT \citep{openai2024chatgpt} are trained to guess the next word in a sequence, which is an intrinsically noisy signal.

Additionally, Cheng \emph{et al.}'s correlation study made some surprising choices:

\begin{itemize}
    \item Cheng \emph{et al.} only report correlation for proxy costs below 0.9 and provide no justification for this decision. This threshold excludes the majority of their own results (e.g. see Cheng \emph{et al.}'s Table 1).
    \item The correlation study considers only a single 45nm test case (Ariane-NG45). NG45 is a much older technology node size and the congestion and density components of the overall cost function should probably be adjusted for better correlation (see Section \ref{benchmarks}).
\end{itemize}

Incidentally, AutoDMP\footnote{AutoDMP is also one of the methods compared against in Cheng \emph{et al.}'s Table 1.} used proxy wirelength, congestion, and density costs similar to those proposed in our \emph{Nature} paper, and found that they do in fact correlate with final metrics \citep{agnesina2023autodmp}.

\subsection{Cheng \emph{et al.}'s Incorrect Claim of Validation by Google Engineers}
\label{false-validation}

Cheng \emph{et al.} claimed that Google engineers confirmed its technical correctness, but this is untrue. Google engineers (who were not corresponding authors of the \emph{Nature} paper) merely confirmed that they were able to train from scratch (i.e. no pre-training) on a single test case from the quick start guide in our open-source repository. The quick start guide is of course not a description of how to fully replicate the methodology described in our \emph{Nature} paper, and is only intended as a first step to confirm that the needed software is installed, that the code has compiled, and that it can successfully run on a single simple test case (Ariane). 

In fact, these Google engineers share our concerns and provided constructive feedback, which was not addressed. For example, prior to publication of Cheng \emph{et al.}, through written communication and in several meetings, they raised concerns about the study, including the use of drastically less compute, and failing to tune proxy cost weights to account for a drastically different technology node size.

The Acknowledgements section of Cheng \emph{et al.} also lists the \emph{Nature} corresponding authors and implies that they were consulted or even involved, but this is not the case. In fact, the corresponding authors only became aware of this paper after its publication.

\section{Transparency \& Reproducibility}
\label{transparency}

\subsection{AlphaChip is Fully Open-Source}
\label{open-source}

We have open-sourced a software repository \citep{guadarrama2021circuittraining} to fully reproduce the methods described in our \emph{Nature} paper. Every line of our RL method is freely available for inspection, execution, or modification, and we provide source code or binaries to perform all preprocessing and postprocessing steps. Open-sourcing the code took over a year of effort by the TF-Agents team\footnote{TensorFlow Agents is a reinforcement learning infrastructure team at Google, which provides open-source libraries. See \url{https://www.tensorflow.org/agents}.}, and included independent replication of our methodology and the results in our \emph{Nature} paper. From our open-source repository: 

\begin{quote}
“Open-sourcing our code involved partnering with another team at Google (TF-Agents). TF-Agents first replicated the results in our \emph{Nature} article using our codebase, then reimplemented our method and replicated our results using their own implementation, and then open-sourced their implementation as it does not rely on any internal infrastructure.”
\end{quote}

Cheng \emph{et al.} unnecessarily ``reverse-engineered'' two functions that we provide as binaries for performance optimization (the proxy cost function and force-directed (FD) standard cell placer). As discussed in an MLCAD 2021 paper \citep{jiang2021delving}, we now recommend using DREAMPlace \citep{yibo2021dreamplace} for standard cell placement, rather than FD, as it yields superior performance. We provide the legacy FD binary for the sole purpose of enabling exact reproduction of our method as published in \emph{Nature}.

Regarding public benchmarks, we reported results on the open-source Ariane RISC-V CPU \citep{zaruba2019ariane} in \emph{Nature}. Additionally, in a follow-up paper at MLCAD 2021 \citep{jiang2021delving}, we evaluated on the open-source ISPD 2015 contest benchmark \citep{bustany2015ispd}. Because we have open-sourced our code, the community is free to follow our methodology and evaluate our method on any public benchmark.

\subsection{Cheng \emph{et al.} Claim They Cannot Share Their ``Open'' Test Cases}
\label{refused}

One of the criticisms put forth in Cheng \emph{et al.} was that the \emph{Nature} evaluation was done on proprietary TPU blocks (in addition to the open-source Ariane block that was also evaluated, and the public ISPD 2015 benchmark in a follow-up publication \citep{jiang2021delving}). Cheng \emph{et al.} claimed to evaluate on a set of open test cases to improve reproducibility, but when we corresponded with the authors, they were unable or unwilling to provide the synthesized netlists necessary to replicate their results on the ``open'' test cases in their main data table (Table 1).

Unfortunately, this means that we cannot replicate any of the results in Cheng \emph{et al.}'s Table 1:

\begin{itemize}
    \item \textbf{GF12 (12nm)}: These test cases are proprietary and unavailable to the public, and Cheng \emph{et al.}'s results are obfuscated, meaning that even if an external researcher were to obtain access, a direct comparison would still be impossible.
    \item \textbf{NG45 (45nm)}: Cheng \emph{et al.} have not shared the synthesized netlists necessary to reproduce their NG45 results, despite 10+ requests since February 2024. Note that other papers evaluate on the NG45 blocks, but their results are inconsistent with those in Cheng \emph{et al.}'s Table 1 (e.g. see Table 2 of AutoDMP \citep{agnesina2023autodmp}), underscoring reproducibility challenges.
\end{itemize} 

It is unfortunate that modern chip IP is sensitive and proprietary, and to our knowledge, there are no open benchmarks available for cutting edge processes. We encourage the chip design community to create more open designs for modern sub-7nm processes, as this will help push the field forward. At the moment, fully open designs are typically 28nm, 45nm, or even 130nm, and many physical design issues are quite different than in sub-7nm processes.

\section{Conclusion}
In Cheng \emph{et al.}'s attempt to reassess our work, the authors did not run our method as described in \emph{Nature} (e.g. they performed no pre-training, used substantially less compute, and did not train to convergence), reported results on benchmarks that are neither representative nor reproducible, and ran questionable ablation/correlation studies.

In his paper \citep{cacm2024}, Markov published baseless allegations of fraud based on a ``meta-analysis'' of Cheng \emph{et al.} (which did not reproduce our methodology) and an anonymous PDF (that Markov actually coauthored), whose results could not be reproduced and for which ``the claims and conclusions in the draft are not scientifically backed by the experiments'' \citep{orwant2022}.

Meanwhile, AlphaChip has inspired an explosion of work on AI for chip design, and its superhuman layouts have been taped out in multiple generations of TPU deployed in Google datacenters all over the world, as well as other chips across Alphabet and by external chipmakers. We look forward to seeing AI continue to transform all aspects of hardware design, just as advances in hardware have revolutionized AI.

\section*{Acknowledgements}

We would like to thank our \emph{Nature} co-authors Mustafa Yazgan, Joe Wenjie Jiang, Ebrahim Songhori, Shen Wang, Young-Joon Lee, Eric Johnson, Omkar Pathak, Azade Nova, Jiwoo Pak, Andy Tong, Kavya Srinivasa, William Hang, Emre Tuncer, Quoc V. Le, James Laudon, Richard Ho, and Roger Carpenter, the TF-Agents team and others involved in contributing to the open-source implementation, especially Sergio Guadarrama, Summer Yue, Toby Boyd, Terence Tam, Guanhang Wu, Kuang-Huei Lee, and Vincent Zhuang, Cliff Young, Dan Belov, Joe Jiang, and Chris Manning for their valuable feedback on this paper, Siddharth Garg, Vijay Reddi, Ruchir Puri, and David Patterson for insightful discussions about the AlphaChip work, and Ed Chi, Zoubin Ghahramani, and Koray Kavukcuoglu for all their advice and support.

\bibliographystyle{plain}
\bibliography{bibliography}

\begin{thebibliography}{10}

\bibitem{ml101}
Charu~C. Aggarwal.
\newblock {\em {Neural Networks and Deep Learning, A Textbook}}.
\newblock Springer International Publishing AG, 2018.

\bibitem{agnesina2023autodmp}
Anthony Agnesina, Puranjay Rajvanshi, Tian Yang, Geraldo Pradipta, Austin Jiao, Ben Keller, Brucek Khailany, and Haoxing Ren.
\newblock {AutoDMP: Automated DREAMPlace-based Macro Placement}.
\newblock {\em {ISPD}}, 2023.

\bibitem{markovetal}
No~authors listed.
\newblock Stronger baselines for evaluating deep reinforcement learning in chip placement, Scan of a document with no publication date.
\newblock https://shorturl.at/DRKP8.

\bibitem{preprint2020}
{Azalia Mirhoseini, Anna Goldie, Mustafa Yazgan, Joe Jiang, Ebrahim Songhori, Shen Wang, Young-Joon Lee, Eric Johnson, Omkar Pathak, Sungmin Bae, Azade Nazi, Jiwoo Pak, Andy Tong, Kavya Srinivasa, William Hang, Emre Tuncer, Anand Babu, Quoc V. Le, James Laudon, Richard Ho, Roger Carpenter, and Jeff Dean}.
\newblock {Chip Placement with Deep Reinforcement Learning}.
\newblock {\em arxiv/2004.10746}, 2020.

\bibitem{yan2024human}
{Binjie Yan, Lin Xu, Zefang Yu, Mingye Xie, Wei Ran, Jingsheng Gao, Yuzhuo Fu, Ting Liu}.
\newblock {Learning to Floorplan like Human Experts via Reinforcement Learning}.
\newblock {\em DATE}, 2024.

\bibitem{bustany2015ispd}
Ismail~S. Bustany, David Chinnery, Joseph~R. Shinnerl, and Vladimir Yutsis.
\newblock {Benchmarks with Fence Regions and Routing Blockages for Detailed-Routing-Driven Placement}.
\newblock {\em {ISPD}}, 2015.

\bibitem{ieee2023}
Rina~Diane Caballar.
\newblock {Q\&A: Here's How AI Will Change Chip Design}.
\newblock {\em {IEEE Spectrum}}, 2023.

\bibitem{cadence2024}
Cadence.
\newblock {Reinforcement Learning}.
\newblock {Available at https://www.cadence.com\\/en\_US/home/explore/reinforcement-learning.html; Accessed: 2024-11-10}.

\bibitem{cheng2023assessment}
Chung-Kuan Cheng, Andrew~B. Kahng, Sayak Kundu, Yucheng Wang, and Zhiang Wang.
\newblock {Assessment of Reinforcement Learning for Macro Placement}.
\newblock {\em ISPD}, 2023.

\bibitem{lai2021maskplace}
Ruoyu Cheng and Junchi Yan.
\newblock {On Joint Learning for Solving Placement and Routing in Chip Design}.
\newblock {\em NeurIPS}, 2021.

\bibitem{chowdhuryrlsynthesis2024}
Animesh~Basak Chowdhury, Marco Romanelli, Benjamin Tan, Ramesh Karri, and Siddharth Garg.
\newblock {Retrieval-Guided Reinforcement Learning for Boolean Circuit Minimization}.
\newblock {\em ICLR}, 2024.

\bibitem{chuang2024aws}
James Chuang and Pedro Gil.
\newblock Boost chip design with {AI}: {How Synopsys DSO.ai on AWS Delivers Lower Power and Faster Time-to-Market}.
\newblock Available at https://aws.amazon.com/blogs/apn/boost-chip-design-with-ai-how-synopsys-dso-ai-on-aws-delivers-lower-power-and-faster-time-to-market; {Accessed: 2024-11-10}.

\bibitem{cheng2019replace}
{Chung-Kuan Cheng, Andrew B. Kahng, Ilgweon Kang, Lutong Wang}.
\newblock {RePlAce: Advancing Solution Quality and Routability Validation in Global Placement}.
\newblock In {\em {IEEE Trans. Comput. Aided Des. Integrated Circ. Syst.}}, 2019.

\bibitem{alphago}
{David Silver, Aja Huang, Chris J. Maddison, Arthur Guez, Laurent Sifre, George van den Driessche, Julian Schrittwieser, Ioannis Antonoglou, Veda Panneershelvam, Marc Lanctot, Sander Dieleman, Dominik Grewe, John Nham, Nal Kalchbrenner, Ilya Sutskever, Timothy Lillicrap, Madeleine Leach, Koray Kavukcuoglu, Thore Graepel, Demis Hassabis}.
\newblock {Mastering the game of Go with deep neural networks and tree search}.
\newblock {\em {Nature}}, 2016.

\bibitem{pan2015}
{David Z. Pan, Lars Liebmann, Bei Yu, Xiaoqing Xu, and Yibo Lin}.
\newblock {Pushing multiple patterning in sub-10nm: Are we ready?}
\newblock {\em DAC}, 2015.

\bibitem{zaruba2019ariane}
{Florian Zaruba and Luca Benini}.
\newblock {The Cost of Application-Class Processing: Energy and Performance Analysis of a Linux-Ready 1.7-GHz 64-Bit RISC-V Core in 22-nm FDSOI Technology}.
\newblock In {\em VLSI Syst.}, 2019.

\bibitem{forbes2023}
Karl Freund.
\newblock {AI Is Reshaping Chip Design. But Where Will It End?}
\newblock Available at {https://www.forbes.com/sites/karlfreund/2023/12/19/ai-is-reshaping-chip-design-but-where-will-it-end/}; {Accessed}: 2024-11-10.

\bibitem{chang2023flexible}
{Fu-Chieh Chang, Yu-Wei Tseng, Ya-Wen Yu, Ssu-Rui Lee, Alexandru Cioba, I-Lun Tseng, Da-shan Shiu, Jhih-Wei Hsu, Cheng-Yuan Wang, Chien-Yi Yang, Ren-Chu Wang, Yao-Wen Chang, Tai-Chen Chen, and Tung-Chieh Chen}.
\newblock {Flexible chip placement via reinforcement learning: late breaking results}.
\newblock {\em DAC}, 2023.

\bibitem{alphachipblog2024}
Anna Goldie and Azalia Mirhoseini.
\newblock {How AlphaChip transformed computer chip design}.
\newblock In {\em Google DeepMind Blog}, 2024.

\bibitem{goldie-mirhoseini2024}
Anna Goldie, Azalia Mirhoseini, Mustafa Yazgan, Joe~Wenjie Jiang, Ebrahim Songhori, Shen Wang, Young-Joon Lee, Eric Johnson, Omkar Pathak, Azade Nazi, Jiwoo Pak, Andy Tong, Kavya Srinivasa, William Hang, Emre Tuncer, Quoc~V. Le, James Laudon, Richard Ho, Roger Carpenter, and Jeff Dean.
\newblock {Addendum: A Graph Placement Methodology for Fast Chip Design}.
\newblock {\em Nature}, 2024.

\bibitem{guadarrama2021circuittraining}
Sergio Guadarrama, Summer Yue, Toby Boyd, Joe~Wenjie Jiang, Ebrahim Songhori, Terence Tam, Anna Goldie, and Azalia Mirhoseini.
\newblock {Circuit Training}: An open-source framework for generating chip floor plans with distributed deep reinforcement learning.
\newblock \url{https://github.com/google_research/circuit_training}, 2021.

\bibitem{jiang2021delving}
Zixuan Jiang, Ebrahim Songhori, Shen Wang, Anna Goldie, Azalia Mirhoseini, Joe Jiang, Young-Joon Lee, and David~Z. Pan.
\newblock {Delving into Macro Placement with Reinforcement Learning}.
\newblock {\em {MLCAD}}, 2021.

\bibitem{karypis1998hmetis}
George Karypis and Vipin Kumar.
\newblock {hMETIS: A Hypergraph Partitioning Package}.
\newblock {\em Manual}, 1998.

\bibitem{laster2022}
Courtney Laster.
\newblock {Declaration of Courtney Laster}.
\newblock {\em Superior Court of California, County of Santa Clara}, 2024.
\newblock Declared Under Penalty of Perjury on June 29, 2022 in Los Angeles, California, Electronically Filed by Superior Court of CA, County of Santa Clara, on 6/30/2022 6:44PM. This document is publicly available. To access it, navigate to https://portal.scscourt.org/search; enter 22CV398683 in the first box entitled `CASE NUMBER SEARCH'; check the I-am-not-a-robot-box (if you are in fact human); click `Search'; click on the middle tab entitled `EVENTS'; navigate to page 12 (the final tab at the bottom); and in the row with the comment `Declaration of Courtney Laster...', click on the PDF icon in the final `Documents' column.

\bibitem{mediatek2022}
Jen-Wei Lee, Yi-Ying Liao, Te-Wei Chen, Yu-Hsiu Lin, Chia-Wei Chen, Chun-Ku Ting, Sheng-Tai Tseng, Ronald Kuo-Hua Ho, Hsin-Chuan Kuo, Chun-Chieh Wang, Ming-Fang Tsai, Chun-Chih Yang, Tai-Lai Tung, and Da-Shan Shiu.
\newblock {A Learning-Based Algorithm for Early Floorplan With Flexible Blocks}.
\newblock In {\em 2022 IEEE Asian Solid-State Circuits Conference}, 2022.

\bibitem{yibo2021dreamplace}
Yibo Lin, Zixuan Jiang, Jiaqi Gu, Wuxi Li, Shounak Dhar, Haoxing Ren, Brucek Khailany, and David~Z. Pan.
\newblock {DREAMPlace: Deep Learning Toolkit-Enabled GPU Acceleration for Modern VLSI Placement}.
\newblock {\em ICCAD}, 2021.

\bibitem{cacm2024}
Igor Markov.
\newblock {Reevaluating Google’s Reinforcement Learning for IC Macro Placement}.
\newblock {\em {Communications of the ACM}}, 2024.

\bibitem{markovarxiv2023}
Igor Markov.
\newblock {The False Dawn: Reevaluating Google’s Reinforcement Learning for IC Macro Placement}.
\newblock {\em {Communications of the ACM}}, 2024.

\bibitem{cadence2022}
Rod Metcalfe.
\newblock {Machine Learning-Driven Full-Flow Chip Design Automation}.
\newblock {\em Cadence}, 2022.

\bibitem{mirhoseini-goldie2021}
Azalia Mirhoseini, Anna Goldie, Mustafa Yazgan, Joe~Wenjie Jiang, Ebrahim Songhori, Shen Wang, Young-Joon Lee, Eric Johnson, Omkar Pathak, Azade Nazi, Jiwoo Pak, Andy Tong, Kavya Srinivasa, William Hang, Emre Tuncer, Quoc~V. Le, James Laudon, Richard Ho, Roger Carpenter, and Jeff Dean.
\newblock {A Graph Placement Methodology for Fast Chip Design}.
\newblock {\em Nature}, 2021.

\bibitem{kahng2022github}
{MacroPlacement Repo. \url{https://github.com/TILOS-AI-Institute/MacroPlacement}}.

\bibitem{openai2024chatgpt}
{OpenAI}.
\newblock {ChatGPT}, 2024.

\bibitem{orwant2022}
Jon Orwant.
\newblock {Declaration of Jon Orwant}.
\newblock {\em Superior Court of California, County of Santa Clara}, 2022.
\newblock Declared Under Penalty of Perjury on June 30, 2022 in Brookline, Massachusetts, Electronically Filed by Superior Court of CA, County of Santa Clara, on 6/30/2022 6:44PM, Publicly Available at https://portal.scscourt.org/, Case Number 22CV398683.

\bibitem{synopsys2024}
Synopsys.
\newblock {What is AI Chip Design?}
\newblock {Available at https://www.synopsys.com/ai/what-is-ai-chip-design.html}; Accessed: 2024-11-10.

\bibitem{physicalsynthesis2024}
Synopsys.
\newblock {What is Physical Synthesis}.
\newblock Available at https://www.synopsys.com/glossary/what-is-physical-synthesis.html; {Accessed}: 2024-11-10.

\bibitem{google2023gemini}
Google~Gemini Team.
\newblock {Gemini: A Family of Highly Capable Multimodal Models}, arXiv:2312.11805, 2023.

\bibitem{eetimes2023}
Sally Ward-Foxton.
\newblock {AI-Powered Chip Design Goes Mainstream}.
\newblock {\em {EE Times}}, 2023.
\newblock Available at https://www.eetimes.com/ai-powered-chip-design-goes-mainstream/; Accessed: 2024-11-10.

\bibitem{multiplepatterning2024}
Wikipedia.
\newblock {Multiple Patterning}, 2024.
\newblock Available at https://en.wikipedia.org/wiki/Multiple\_patterning; {Accessed}: 2024-04-23.

\bibitem{lai2023chipformer}
{Yao Lai, Jinxin Liu, Zhentao Tang, Bin Wang, Jianye Hao, Ping Luo}.
\newblock {ChiPFormer: Transferable Chip Placement via Offline Decision Transformer}.
\newblock {\em {PMLR}}, 2023.

\bibitem{lai2022maskplace}
{Yao Lai, Mu Yao, Luo Ping}.
\newblock {MaskPlace: Fast Chip Placement via Reinforced Visual Representation Learning}.
\newblock {\em NeurIPS}, 2022.

\bibitem{lu2023dac}
{Yi-Chen Lu, Wei-Ting Chan, Deyuan Guo, Sudipto Kundu, Vishal Khandelwal, and Sung Kyu Lim}.
\newblock {RL-CCD: Concurrent Clock and Data Optimization using Attention-Based Self-Supervised Reinforcement Learning}.
\newblock {\em {DAC}}, 2023.

\bibitem{yue2022circuittraining}
Summer Yue, Ebrahim~M. Songhori, Joe~Wenjie Jiang, Toby Boyd, Anna Goldie, Azalia Mirhoseini, and Sergio Guadarrama.
\newblock {Scalability and Generalization of Circuit Training for Chip Floorplanning}.
\newblock {\em {ISPD}}, 2022.

\bibitem{shi2024wiremask}
{Yunqi Shi, Ke Xue, Lei Song, Chao Qian}.
\newblock {Macro Placement by Wire-Mask-Guided Black-Box Optimization}.
\newblock {\em NeurIPS}, 2024.

\bibitem{ziemelis2024}
Karl Ziemelis.
\newblock \emph{Nature} {Chief Physical Sciences Editor}, {April 12, 2024}.
\newblock {\em Personal communication}, 2024.

\end{thebibliography}

\end{document}